%% file: root.tex
\documentclass[letterpaper, 10 pt, journal, twoside]{IEEEtran}
\IEEEoverridecommandlockouts
\usepackage{hyperref}
\hypersetup{pdftex,colorlinks=true,allcolors=blue}
\usepackage{hypcap}
\usepackage{amsmath} 
\usepackage{amssymb}  
\usepackage{graphicx}
\usepackage{caption}
\usepackage{subcaption}
\usepackage{multirow}
\usepackage{amsfonts}
\usepackage{tabularx}
\usepackage{amsfonts}
\usepackage{algorithm,algpseudocode}
\usepackage{bm}
\usepackage{enumerate}
\usepackage{verbatim}
\usepackage{float}
\usepackage{siunitx}
\usepackage{mdframed}
\usepackage{adjustbox}
\usepackage{cleveref}
\usepackage{authblk}
\usepackage{color}
\usepackage{verbatim}
\usepackage{booktabs}
\usepackage{comment}

\title{\LARGE \bf Multi-Robot Collaborative Perception with Graph Neural Networks}
\author{Yang Zhou, Jiuhong Xiao, Yue Zhou, and Giuseppe Loianno
\thanks{Manuscript received: September, 9, 2021; Revised November, 23, 2021; Accepted December, 20, 2021.}
\thanks{This paper was recommended for publication by Editor Cesar Cadena upon evaluation of the Associate Editor and Reviewers' comments. This work was supported by Qualcomm Research, the NSF CPS Grant CNS-2121391, the Technology Innovation Institute, Nokia, and the NYU Wireless.}
\thanks{The authors are with the New York University, Brooklyn, NY 11201, USA. {\tt\footnotesize email: \{yangzhou, jx1190, yz1268, loiannog\}@nyu.edu}.}
\thanks{We thank Jatin Palchuri and Rundong Ge for help on dataset collection.}
\thanks{Digital Object Identifier (DOI) 10.1109/LRA.2022.3141661}
}
\begin{document}
\IEEEpubid{\begin{minipage}[t]{\textwidth}\ \\[10pt]        {\copyright 2022 IEEE. Personal use is permitted, but republication/redistribution requires IEEE permission. See https://www.ieee.org/publications/rights/index.html for more information.}\end{minipage}} 
\maketitle

\begin{abstract}
Multi-robot systems such as swarms of aerial robots are naturally suited to offer additional flexibility, resilience, and robustness in several tasks compared to a single robot by enabling cooperation among the agents. To enhance the autonomous robot decision-making process and situational awareness, multi-robot systems have to coordinate their perception capabilities to collect, share, and fuse environment information among the agents efficiently to obtain context-appropriate information or gain resilience to sensor noise or failures. In this paper, we propose a general-purpose Graph Neural Network (GNN) with the main goal to increase, in multi-robot perception tasks, single robots' inference perception accuracy as well as resilience to sensor failures and disturbances. We show that the proposed framework can address multi-view visual perception problems such as monocular depth estimation and semantic segmentation. Several experiments both using photo-realistic and real data gathered from multiple aerial robots' viewpoints show the effectiveness of the proposed approach in challenging inference conditions including images corrupted by heavy noise and camera occlusions or failures. 
\end{abstract}
\section{Introduction}~\label{sec:introduction}
\input{chap/1_introduction.tex}
\section{Related Works}~\label{sec:relatedworks}
\input{chap/2_related_works.tex}
\section{Methodology}~\label{sec:methodology}
\input{chap/3_methods.tex}
\section{Multi-view Visual Perception Case Study}~\label{sec:multi-viewperception}
\input{chap/3a_multi_view_perception_case.tex}
\section{Experiments}~\label{sec:exepriements}
\input{chap/4_experiments.tex}
\section{Conclusion}~\label{sec:conclusion}
\input{chap/5_conclusion.tex}
%\vspace{-10pt}
\bibliographystyle{IEEEtran} % use IEEEtran.bst style
\bibliography{mybib}
\end{document}

%% file: chap/1_introduction.tex
\IEEEPARstart{I}{n} the past decade, robot perception played a fundamental role in enhancing robot autonomous decision-making processes in a variety of tasks including search and rescue, transportation, inspection, and situational awareness. Compared to a single robot, the use of multiple robots can speed up the execution of time-critical tasks while concurrently providing resilience to sensor noises and failures.

Multi-robot perception includes a wide range of collaborative perception tasks with multiple robots (e.g., swarms of aerial or ground robots) such as multi-robot detection, multi-robot tracking, multi-robot localization and mapping. They aim to achieve accurate, consistent, and robust environment sensing, which is crucial for collaborative high-level decision-making policies and downstream tasks such as planning and control. There are still several major challenges in multi-robot perception. Robots must efficiently collaborate with each others with  minimal or even absence of communication.
This translates at the perception level in the ability to collect and fuse information from the environments and neighbour agents in an efficient and meaningful way to be resilient to sensor noise or failures. Information sharing is fundamental to accurately obtain context-appropriate information and provides resilience to sensor noise, occlusions, and failures. These problems are often resolved by leveraging expert domain knowledge to design specific and handcrafted mechanisms for different multi-robot perception problems. 

In this work, we address the multi-robot perception problem with GNNs. 
Recent trend of deep learning has produced a paradigm shift in multiple fields including robotics. Data-driven approaches~\cite{garg2020Semantics} have outperformed classical methods in multiple robot perception problems, including monocular depth estimation, semantic segmentation, object detection, and object tracking without requiring expert domain knowledge. The single-robot system can benefit from the thrive of deep neural networks and collaboration with other agents.
\begin{figure}[t]
    \centering
    \includegraphics[width=0.9\linewidth]{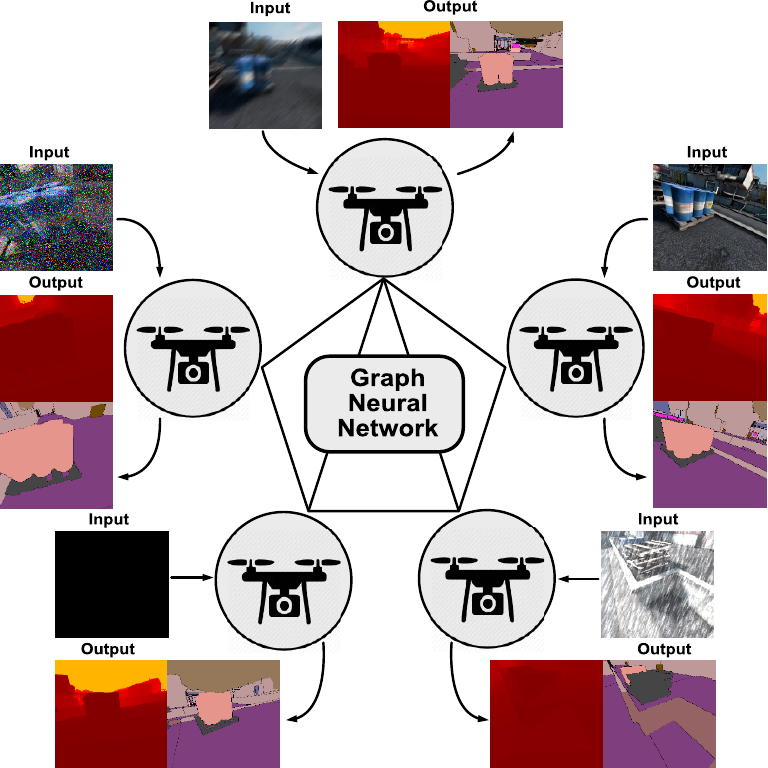}
    \caption{Multi-robot collaborative perception with GNN.
    In this case, images are the inputs and monocular depth estimation and semantic segmentation are the downstream tasks.
    \label{fig:teaser}}
     \vspace{-20pt}
\end{figure}
Graph Neural Networks (GNNs) can exploit the underlying graph structure of the multi-robot perception problem by utilizing the message passing among nodes in the graph. The node features are updated by one or multiple rounds by aggregating node features from the neighbors. Various types of graph neural networks have been proposed, including GNN~\cite{scarselli2009Graph}, Convolutional GNNs~\cite{kipf2017SemiSupervised}, Graph Attention Networks~\cite{velickovic2018Graph}. These methods have proved to be effective for node classification, graph classification, and link prediction. Recently, researchers also started to apply GNNs on multi-robot systems for communication~\cite{li2020Graph}  and planning~\cite{tolstaya2020Learning}. However, there is still no work exploiting the expressiveness of GNNs for multi-robot perception problems.

This work presents multiple contributions. First, we propose a generalizable GNN-based perception framework for multi-robot systems to increase single robots' inference perception accuracy, which is illustrated in Fig.~\ref{fig:teaser}. Our approach is flexible and takes into account different sensor modalities. It embeds the spatial relationship between neighbor nodes into the messages and employs the cross attention mechanism to adjust message weights according to the correlation of node features of different robots. Second, we show the proposed approach in two multi-view visual perception tasks: collaborative monocular depth estimation and semantic segmentation and we discuss how to employ the proposed framework with different sensing modalities.  
Finally, we show the effectiveness of the proposed approach in challenging multi-view perception experiments including camera sensors affected by heavy image noise, occlusions, and failures on photo-realistic as well as real image data collected with aerial robots. To the best of our knowledge, this is the first time that a GNN has been employed to solve a multi-view/multi-robot perception task using real robot image data.

The paper is organized as follows. Section~\ref{sec:relatedworks} presents a literature review. Section~\ref{sec:methodology} presents our GNN architecture whereas Section~\ref{sec:multi-viewperception} shows two instances of our method to address in two multi-view visual perception problems. In Section~\ref{sec:exepriements}, several experiments are presented to validate the proposed approach. Section~\ref{sec:conclusion} concludes the paper.

%% file: chap/2_related_works.tex
Single-robot scene understanding benefited from the expressive power of deep neural networks. Among these frame-based perception tasks, semantic segmentation~\cite{long2015Fully, ronneberger2015UNet, chen2018DeepLab} and monocular depth estimation~\cite{eigen2014Depth,  godard2019Digging} are among important perception problems which have been widely studied in computer vision and robotics community. For both tasks, researchers generally adopt an encoder-decoder structure deep neural network. \cite{long2015Fully} first employs a fully convolutional network for semantic segmentation. UNet~\cite{ronneberger2015UNet} introduces the skip connection, and Deeplab~\cite{chen2018DeepLab} focuses on improving the decoder of the network architecture in order to exploit the feature map extracted from it. In~\cite{eigen2014Depth}, the authors also use deep neural networks to solve monocular depth estimation. Most recently, unsupervised learning methods such as  Monodepth2~\cite{godard2019Digging} still follow the encoder-decoder network structure for monocular depth estimation.

Many tasks of interest for the robotics and computer vision communities such as 3D shape recognition~\cite{su2015MultiView, wei2020ViewGCN}, object-level scene understanding~\cite{lin2021Multiview}, and object pose estimation~\cite{labbe2020CosyPose} can benefit from using multi-view perception. This can improve the task accuracy as well as increase robustness with respect to single robot sensor failures. Multi-robot perception also includes the problems related to communication and bandwidth limitation among the robot network. Who2com~\cite{liu2020Who2com} proposes a multi-stage handshake communication mechanism that enables agents to share information with limited communication bandwidth. The degraded agent only connects with the selected agent to receive information. When2com~\cite{liu2020When2com} as its following work further investigates the multi-agent communication problem by constructing communication groups and deciding when to communicate based on the communication bandwidth. They validated their approach on semantic segmentation with noisy sensor inputs. 
Conversely, in our work, we study the multi-robot perception problem employing a general GNN-based framework which directly operates on the graph structure. The proposed framework enables to resolve the multi-robot perception problem by leveraging the representation power benefits of modern GNN methods. We experimentally demonstrate the robustness to heavy exogenous (sensor and environment) noise or sensor corruption.

GNNs can model problems with structures of directional and un-directional graphs. The graph neural network propagates and aggregates messages around neighbor nodes to update each node feature~\cite{scarselli2009Graph}. In addition to nodes, the edges can also embed features~\cite{schlichtkrull2018Modeling}. Researchers focus on various approaches to create individual node embeddings. GraphSAGE~\cite{hamilton2017Inductive} learns a function to obtain embedding by sampling and aggregation from neighbor nodes. Graph Attention Networks~\cite{velickovic2018Graph} introduces a multi-head attention mechanism to dynamically adjust weights of neighbor nodes in the aggregation step. MAGAT~\cite{li2021message} introduces a key-query-like attention mechanism to improve the communication efficiency. GNNs have been applied in different fields of robotics \cite{li2020Graph, tolstaya2020Learning,li2021message,khan2020Graph, kortvelesy2021ModGNN} including multi-robot path planning, decentralized control, and policy approximation. These works show the potential of GNNs in the multi-robot system by modeling each robot as a node in the graph. Leveraging GNN, the method can learn efficient coordination in the multi-robot system and can be generalized to different graph topology with shared model parameters. 
There are few perception-related works exploiting the power of GNNs. \cite{xue2020Learning} studies the multi-view camera relocalization problem whereas Pose-GNN~\cite{elmoogy2021PoseGNN} proposes an image-based localization approach. In these works, node features are initialized from image inputs by convolutional neural network encoders. Both methods treat their problems as supervised graph node regression problems. 

%% file: chap/3_methods.tex
\vspace{-30pt}
\subsection{Preliminaries}~\label{sec:preliminaries}
\begin{figure*}[ht]
    \centering
    \includegraphics[width=0.9\linewidth]{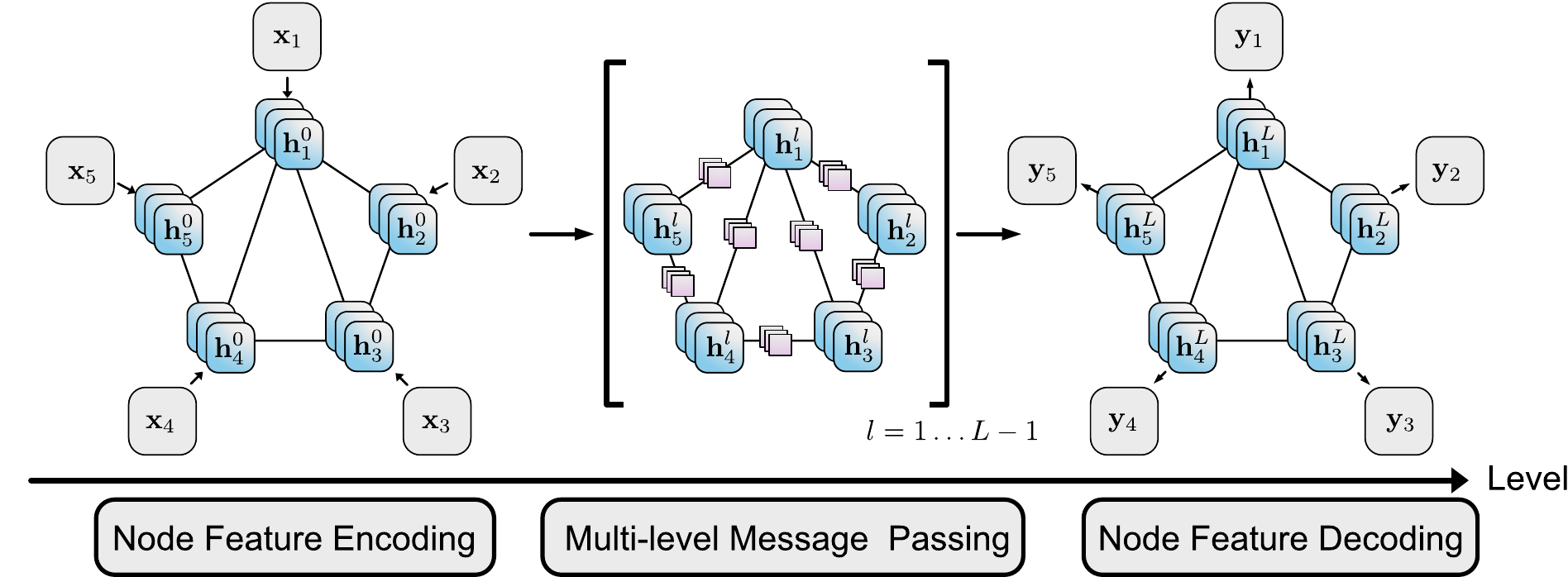}
    \caption{System Overview of the proposed GNN.}
    \label{fig:system_overview}
     \vspace{-20pt}
\end{figure*}
We denote with $N$ the number of robots in a multi-robot perception system. We consider each robot $i$ as a node $v_i$. If robot $i$ and robot $j$ have communication connection, we construct an edge $e_{ij}$ between $v_i$ and $v_j$. Hence, we construct a graph $\mathcal{G}=(\mathcal{V},\mathcal{E}),~\mathcal{V}=\{v_i\},~\mathcal{E}=\{e_{ij}\}, i,j\in\{1,\dots, N\}$, according to the communication connection between robots. The communication topology can be determined by considering a distance threshold between robots or the strength of the communication signal.
The proposed multi-robot perception system, illustrated in Fig.~\ref{fig:system_overview}, takes observations $\{\mathbf{x}_i\}_{i=1\dots N}$ of sensors from all robots, and returns the output $\{\mathbf{y}_i\}_{i=1\dots N}$ after the GNN processing.
We denote the neighbor nodes of $v_i$ as $\mathcal{N}_{(i)}$.
We assume the graph structure does not change between the time that sensors capture information and the time that the GNN provides the results. 
The GNN first encodes the observation $\{\mathbf{x}_{i}\}_{i=1\dots N}$ into the node feature $\{\mathbf{h}_{i}\}_{i=1\dots N}$  where $\mathbf{h}_i ^0 = f_{\mathrm{Encode}} (\mathbf{x}_i)$.
At each level $l \in{1\dots L}$ of the message passing, each node $v_i$ aggregates messages $\mathbf{m}_{ij}^{l}$ from its neighbor nodes $\{v_{j}\}_{j\in\mathcal{N}_{(i)}}$. We consider two different message encoding mechanisms: a) spatial encoding and b) dynamic cross attention encoding. These mechanisms make the messages on bidirectional edges of the graph nonidentical.
After each level of the message passing, the node feature $\mathbf{h}_i^l$ is updated. Once the last level $L$ of message passing is executed, the final result $\mathbf{y}_i$ is obtained by decoding the node feature $\mathbf{h}_i^L$.

In the following, we detail our proposed method.
We explain how to use the determined spatial encoding to utilize the relative spatial relationship between two nodes in Section~\ref{sec:determined_spatial_encoding}. We also show the way to encode feature map correlation into messages by cross attention in Section~\ref{sec:cross_attention_encoding}. The message passing mechanism is introduced in Section~\ref{sec:message_passing} whereas the design of feature decoder in Section~\ref{sec:feature_decoder}.

\vspace{-10pt}
\subsection{Messages with Spatial Encoding}\label{sec:determined_spatial_encoding}

\begin{figure}[!b]
    \centering
    \includegraphics[width=0.9\linewidth]{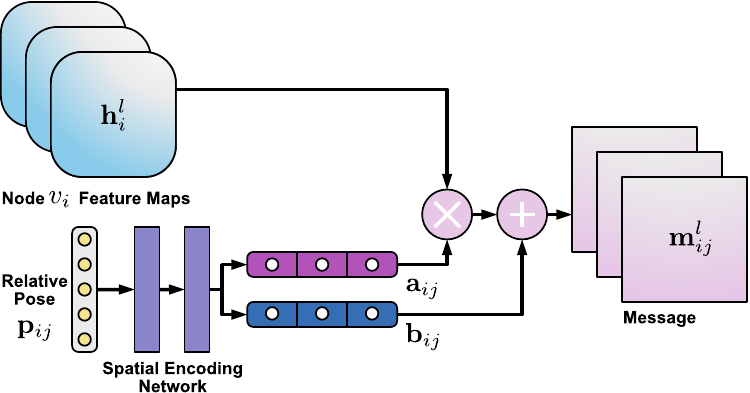}
    \caption{Message with spatial encoding.}
    \label{fig:pose-encoding}
     \vspace{-10pt}
\end{figure}
For robotics application, relative spatial relationship of the multi-robot network can be obtained by the multi-robot system itself~\cite{paliv2021Tracking} or from other systems~\cite{guler2019Infrastructure}. If we have access to the relative spatial relationship between nodes, we can encode it into the messages shared between the node pairs. We illustrate the message generation mechanism with spatial encoding in Fig.~\ref{fig:pose-encoding}.
We can represent the relative pose between robot $i$ and robot $j$ which contains rotation $\mathbf{R}_{ij}$ and translation $\mathbf{t}_{ij} = [t^x_{ij}, t^y_{ij}, t^z_{ij}]$. In \cite{zhou2019Continuity}, a continuous rotation representation $\mathbf{R}^{'}_{ij}$ which takes first two columns of $\mathbf{R}_{ij}$ 
\begin{align}
    \mathbf{R}_{ij} &= 
    \begin{bmatrix}
        r_1 & r_4 & r_7\\
        r_2 & r_5 & r_8\\
        r_3 & r_6 & r_9
        \end{bmatrix},~
    \mathbf{R}^{'}_{ij} =
    \begin{bmatrix}
        r_1 & r_4 \\
        r_2 & r_5\\
        r_3 & r_6
        \end{bmatrix},
    \vspace{-10pt}
\end{align}
\begin{align}
    \mathbf{p}_{ij} &= [t^x_{ij}, t^y_{ij}, t^z_{ij},  r_1 , r_2, r_3, r_4, r_5, r_6].
    \vspace{-10pt}
\end{align}
Compared to common rotation representations including rotation matrix, quaternion and axis angles, this continuous rotation representation is easier to learn for neural networks~\cite{zhou2019Continuity}. We feed the continuous relative pose representation $\mathbf{p}_{ij}$ into the spatial encoding network to produce $\{\mathbf{a}_{ij}, \mathbf{b}_{ij}\}=\{{(a_{ij})}_k,{(b_{ij})}_k\}_{k=1,\dots, C},~{(a_{ij})}_k\in \mathbb{R},~{(b_{ij})}_k\in\mathbb{R}$ where $C$ is number of channel of the node feature $\mathbf{h}_i^l$. For simplicity, we implement the spatial encoding network as two fully connected layers with $\mathrm{ReLU}$. Inspired by FiLM~\cite{perez2018FiLM}, we embed the relative robot pose $\mathbf{p}_{ij}$ and the node feature $\mathbf{h}_i^l$ to compose the message from the node $v_i$ to the node $v_j$. The $k$-th channel of the message transforms $\mathbf{m}_{ij}$ as
\begin{align}
    (\mathbf{a}_{ij}, \mathbf{b}_{ij}) &= \operatorname{FiLM}(\mathbf{p}_{ij}), \\
    {(\mathbf{m}_{ij}^{l})}_k &= {(a_{ij})}_k {(\mathbf{h}_i^l)}_k  + {(b_{ij})}_k \mathbf{1},
\end{align}
where $\mathbf{1}$ is a matrix of ones with the same dimension as the node feature. $\{{(a_{ij})}_k,{(b_{ij})}_k\}_{k=1,\dots, C}$ serves as a set of affine transformation parameter applied on the node feature $\mathbf{m}_{ij}$. It embeds the relative spatial relationship between robot $i$ and robot $j$ onto the node feature $\mathbf{h}_{i}^{l}$ to compose the message $\mathbf{m}_{ij}^{l}$ shared from the node $v_{i}$ to the node $v_{j}$ in the GNN.
\vspace{-10pt}
\subsection{Messages with Dynamic Cross Attention Encoding}\label{sec:cross_attention_encoding}
\begin{figure}[!t]
    \centering
    \includegraphics[width=0.6\linewidth]{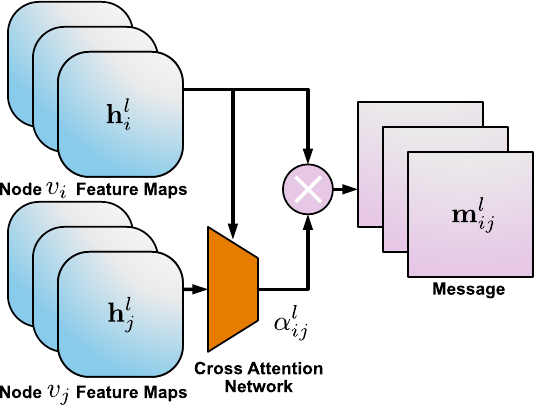}
    \caption{Message passing with cross attention.}
    \label{fig:attention-encoding}
    \vspace{-20pt}
\end{figure}
We also propose another message encoding mechanism with dynamic cross attention between different node features inspired by Graph Attention Networks~\cite{velickovic2018Graph}. This message encoding mechanism takes the dynamic feature relationship of neighbor robots into account and weights the messages from neighbor nodes accordingly.
We illustrate this message encoding mechanism in Fig.~\ref{fig:attention-encoding}. Consider the node $v_i$ and the node $v_j$, we take cross attention of the node features $\mathbf{h}_i^l$ and $\mathbf{h}_j^l$ to obtain a single scaled value as the attention weight of node feature $\mathbf{h}_i^l$.
We first concatenate both feature maps $\mathbf{h}_{i}^{l},~\mathbf{h}_{j}^{l}$, and transform the concatenated feature map to $\alpha_{ij}^{l}\in\mathbb{R}$ by a linear transform module $\mathrm{F}(\cdot)$ and a nonlinear layer $\mathrm{LeakyReLU}(\cdot)$. The output of the transform module is a single scaled value.
We use $\mathrm{Softmax}$ to normalize the attention score $\alpha_{ij}^{l}$ on the incoming edges $e_{ij}$ of the node $v_{j}$ in order to make attention scores more comparable across different neighbor nodes as
\begin{equation}
\begin{split}
    \mathbf{m}_{ij}^{l} &= \mathrm{Softmax}_{i} (\alpha_{ij}^{l}) \mathbf{h}_{i}^{l}, \\
    \alpha_{ij}^{l} &= \mathrm{LeakyReLU}\left(\mathrm{F} [  \mathbf{h}_{i}^l \|  \mathbf{h}_{j}^l]\right).
\end{split}
\end{equation}

In order to enrich the model capacity, we also introduce a multi-head attention mechanism. We use $D$ different transform layer $\mathrm{F^d}$ instead of a single-head attention mechanism to embed the concatenated node feature $[\mathbf{h}_{i}^l \|  \mathbf{h}_{j}^l]$. Each attention head generates a message ${(\mathbf{m}_{ij}^l)}^{d}$. The final message is the average of the $D$ messages
\begin{align}
    \mathbf{m}_{ij}^{l} &= \frac{1}{D} \sum_{d=1}^D {(\mathbf{m}_{ij}^l)}^{d} ,\\ 
    {(\mathbf{m}_{ij}^l)}^{d} &= \mathrm{Softmax}_{i} ({(\alpha_{ij}^{l})}^{d}) \mathbf{\mathbf{h}}_{i}^{l}, \\
    {(\alpha_{ij}^l)}^{d} &= \mathrm{LeakyReLU}\left(\mathrm{F^d} [  \mathbf{h}_{i}^l \| \mathbf{h}_{j}^l]\right).
    \vspace{-10pt}
\end{align}
\vspace{-20pt}
\subsection{Message Passing Mechanism}\label{sec:message_passing}
The message passing mechanism aggregates messages $\mathbf{m}_{ji}$ among neighbor nodes $v_j$ and $v_i$, and update the node next level feature $\mathbf{h}_{i}^{l+1}$. We simply use an average operation as the aggregation operation instead of other options with extra parameters to keep computation and memory consumption suitable for real-time robotics application
\begin{align}
    \mathbf{h}_{i}^{l+1} =  \frac{1}{\# \mathcal{N}(i)} \sum_{j\in\mathcal{N}(i)} \mathbf{m}_{ji}^{l}.
    \vspace{-20pt}
\end{align}
\vspace{-20pt}
\subsection{Feature Decoder} \label{sec:feature_decoder}
\begin{figure}[!t]
    \centering
    \includegraphics[width=0.8\linewidth]{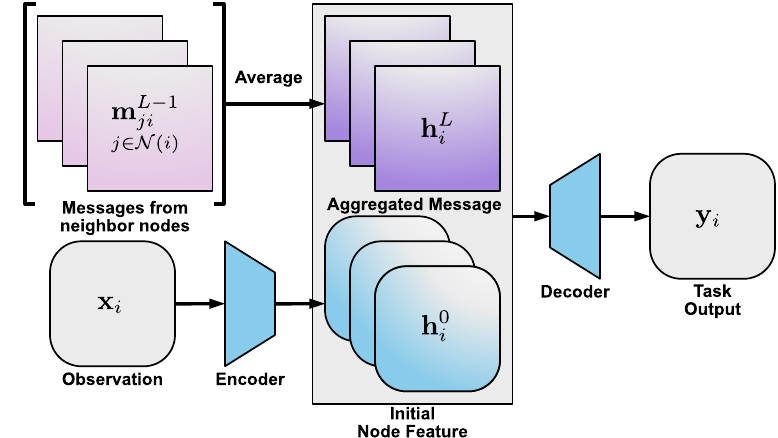}
    \caption{Node feature maps aggregation and feature decoder.}
    \label{fig:feature_decoder}
     \vspace{-20pt}
\end{figure}

We introduce the feature decoding mechanism in Fig.~\ref{fig:feature_decoder}.
After the final levels of message passing, the node feature of each node has aggregated the information from other related robots according to spatial relationship by determined spatial message encoding or feature relationship by dynamic cross attention message encoding. We concatenate the original node feature $\mathbf{h}_i^0$ with the last updated node feature $\mathbf{h}_i^L$ without losing the original encoded information as
\begin{equation}
    \mathbf{y}_i = f_{\mathrm{Decode}} [\mathbf{h}_i^0 \| \mathbf{h}_i^L].
    \vspace{-3pt}
\end{equation}
\vspace{-10pt}
\subsection{Task-related loss}
We train our entire framework with supervised learning. The task-related loss is designed as the mean of objective functions of all robots. The objective function $F_{\text{loss}}(\cdot)$ can be designed according to the specific sensor modalities and tasks. In Section~\ref{sec:multi-viewperception}, we show the design for the multi-view visual perception case. The term $\mathbf{y}_i$ is the prediction obtained from the proposed method, $\mathbf{y}_i^*$ is the target ground truth.
\begin{align}
    L = \frac{1}{N} \sum_{i=1}^N F_{\text{loss}} (\mathbf{x}_i, \mathbf{y}_i, \mathbf{y}_i^*).
    \vspace{-3pt}
\end{align}

%% file: chap/3a_multi_view_perception_case.tex
Our framework is generalizable to different sensor modalities. In order to show its effectiveness, we demonstrate the proposed approach in two multi-view visual perception problems, semantic segmentation and monocular depth estimation. We leverage Deep Graph Library (DGL)~\cite{wang2020Deep} Python package and PyTorch~\cite{paszke2019PyTorch} to design our approach.

We take 2D images $\mathbf{I}\in \mathbb{R}^{H\times W}$ as sensor observations $\mathbf{x}_i$. Therefore, we use a differentiable 2D Convolutional Neural Network (CNN) as feature encoder. In order to meet the real-time requirement of the embedded robotic application, we employ MobileNetV2~\cite{sandler2018MobileNetV2} as a lightweight network for feature encoding. The dimension of the node feature $h_i^l$ is $C\times h\times w$, where $(h,w)$ are the scaled dimension of original image size $(H,W)$. 
The message encoding presented in Section~\ref{sec:determined_spatial_encoding} and Section~\ref{sec:cross_attention_encoding} uses a 2D convolution operator and matches the dimension of the node feature.
The $k$-th channel of the message transforms $\mathbf{m}_{ij}$ as 
\begin{align}
    {(\mathbf{m}_{ij}^{l})}_k = {(a_{ij}^{l})}_k {(\mathbf{h}_i^l)}_k  + {(b_{ij}^{l})}_k \mathbf{1}^{h\times w},
\end{align}
where $\mathbf{1}^{h\times w}$ is a matrix of ones with the same dimension as the node feature $h_i^l\in\mathbb{R}^{C\times h\times w}$ when the input modality is a 2D image.
The transform module $\mathrm{F}(\cdot)$ described in Section~\ref{sec:cross_attention_encoding} is based on two 2D convolution layers to reduce the dimension of the feature map to $1/8$ of the original size, then it flattens it to 1D feature vector. Subsequently, the transform module applies a fully connected layer to transform the 1D feature vector to a single scaled value.

At the feature decoder stage introduced in Section~\ref{sec:feature_decoder}, in order to recover the concatenated feature map to the original size, we use 2D transposed convolution with $\mathrm{ReLU}(\cdot)$ as a nonlinear activation function to enlarge the size of the  feature map through $5$ layers. We keep the design of the decoder simple compatible with our case studies: semantic segmentation and monocular depth estimation. 

As a task-related loss, we employ objective functions which are widely used for semantic segmentation~\cite{long2015Fully} and monocular depth estimation~\cite{godard2019Digging}.
For monocular depth estimation, we use smooth L1 loss and edge-aware smooth loss
\begin{equation}
\begin{split}
    F_{depth}(\mathbf{x}_i, \mathbf{y}_i, \mathbf{y}_i^*) =& L_{L1} (\mathbf{y}_i, \mathbf{y}_i^*) + \alpha L_{s} (\mathbf{y}_i, \mathbf{x}_i),\\
    L_{L1} (\mathbf{y}_i,\mathbf{y}_i^*)=& \begin{cases}0.5\left(\mathbf{y}_{i}-\mathbf{y}_{i}^*\right)^{2} /\beta, & \hspace{-5pt}\text {if }\left|\mathbf{y}_{i}-\mathbf{y}_{i}^*\right|<\beta \\ \left|\mathbf{y}_{i}-\mathbf{y}_{i}^*\right|-0.5 \beta, & \hspace{-5pt}\text {otherwise}
    \end{cases},\\
    L_{s}(\mathbf{y}_i,\mathbf{x}_i) =& \left|\partial_{x} \mathbf{y}_i\right| e^{-\left|\partial_{x} \mathbf{x}_{i}\right|}+\left|\partial_{y} \mathbf{y}_i\right| e^{-\left|\partial_{y} \mathbf{x}_i\right|}.
\end{split}
\end{equation}
where $\partial_x (\cdot), \partial_y(\cdot)$ represent the gradient operations that apply on 2D depth prediction $\mathbf{y}_i$ and original 2D image $\mathbf{x}_i$. The edge-aware loss encourages the gradient of the depth prediction to be consistent with the original image. 

Conversely, for semantic segmentation, we use a cross entropy loss function
\begin{equation}
    F_{semantic}(\mathbf{y}_i, \mathbf{y}_i^*) = L_{CE} (\mathbf{y}_i, \mathbf{y}_i^*).
\end{equation}

Our method can also be easily employed with different sensor modalities by changing the dimension of node features and the design of encoder and decoder structures. For example, it is possible to use a recurrent neural network (RNN) to encode IMU measurements or a 3D convolutional neural network to encode 3D LiDAR measurements. The user also needs to adapt the task-related loss for other tasks. Takes visual object pose estimation for example, the user should use the distance between points on the ground truth pose and the estimated pose as the task-related loss.

%% file: chap/4_experiments.tex
\vspace{-30pt}
\subsection{Dataset}
We collected photo-realistic simulation and real-world multi-robot data. 
We simulate different categories of sensor noises similarly to~\cite{hendrycks2019Benchmarking}.  We consider four datasets with different properties to validate our approach in different scenarios.  All simulated datasets contain depth and semantic segmentation ground truth. The real-world dataset contains depth ground truth. In the four datasets, we simulate the sensor corruption or sensor noise by 
corrupting images and varying the number of affected cameras as described below.
\subsubsection{Photo-realistic Data}
\textbf{Airsim-MAP} dataset was first proposed in When2com~\cite{liu2020When2com}. It contains images from five robots flying in a group in the city environment and rotating along their own z-axis with different angular velocities. In the noisy version of the dataset, \textbf{Airsim-MAP} applies Gaussian blurring with random kernel size from $1$ to $100$ and Gaussian noise \cite{liu2020Who2com}. Each camera has over $50\%$ probability of being corrupted and each frame has at least one noisy image in five views~\cite{liu2020When2com}. We use this dataset to show that our approach is robust to sensor corruption. 

\textbf{Industrial-pose}, \textbf{Industrial-circle} and \textbf{Industrial-rotation} datasets are generated leveraging the Flightmare~\cite{song2020Flightmare} simulator. We use five flying robots in an industrial harbor with various buildings and containers. In all the datasets, the robots form a circle. 
%in the x-y plane. 
\textbf{Industrial-pose} dataset has more variations in terms of relative pose combination among the vehicles and the robots fly at a higher altitude compared to the other two datasets.  In the \textbf{Industrial-circle} dataset, the overlapping region sizes across the robots' Field of View (FoV) are larger compared to the \textbf{Industrial-rotation} dataset.
Specifically, robots in the \textbf{Industrial-circle} are pointing to the center of the team whereas robots in the \textbf{Industrial-rotation} are pointing out of the circle. We project the robots' FoV to the 2D ground plane and calculate the ratio between the union and the sum of the projected FoVs, where the ratio of \textbf{Industrial-circle} is $24.4\%$ and the ratio of  \textbf{Industrial-rotation} is $18.4\%$. Industrial datasets use \cite{hendrycks2019Benchmarking} to generate noise. The noise types include motion blur, Gaussian noise, shot noise, impulse noise, snow effects, and JPEG compression effects. These datasets select the first $N(0-2)$ cameras to perturb. Compared to \textbf{Airsim-MAP}, noise on industrial datasets is less severe, but presents a larger variety.
We use \textbf{Industrial-pose} dataset to show that our approach is robust to different sensor noises and the \textbf{Industrial-circle} and \textbf{Industrial-rotation} to verify the robustness of the proposed method to different overlap areas between robots' FoV.

\subsubsection{Real Data}
\begin{figure}[]
    \centering
    \includegraphics[width=0.8\linewidth]{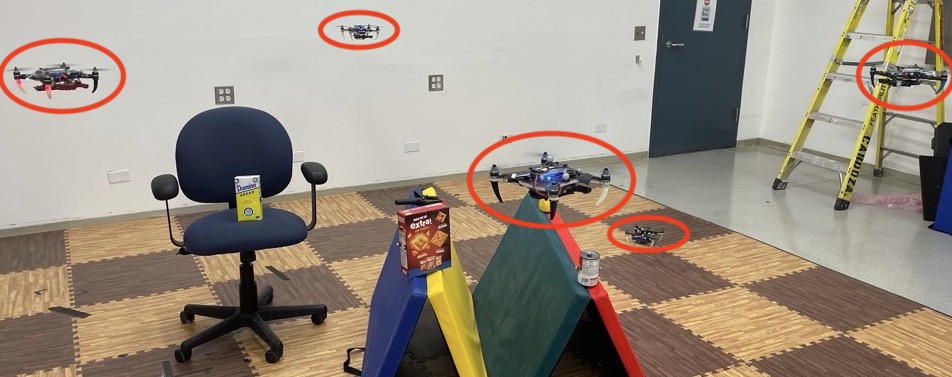}
    \caption{Real data collected by five aerial robots with cameras.}
    \label{fig:real-robot}
    \vspace{-20pt}
\end{figure}

\textbf{ARPL-r4} dataset was collected in an indoor flying arena of $10\times6\times4~\si{m^3}$ at the Agile Robotics and Perception Lab (ARPL) lab at New York University using multiple aerial robots based on our previous work~\cite{LoiannoRAL2017} equipped with RGB camera that autonomously fly and observe a feature-rich object (see Fig.~\ref{fig:real-robot}). We use this dataset to validate our method in real-world scenarios. The vehicles are performing extreme maneuvers creating motion blur. The robots form a circle and point their front camera towards the object in the center to maximize their overlapping visual area. We simulate different noise patterns on one or two robots.

\vspace{-5pt}
\subsection{Evaluation Protocol}
In our experiments, we compare the performance of different methods on the monocular depth estimation and semantic segmentation. We use Abs Relative difference (Abs Rel), Squared Relative difference (Sq Rel) and Root Mean Squared Error (RMSE) as the metric for monocular depth estimation~\cite{eigen2014Depth}. We use mIoU as the metric of semantic segmentation. The bolded numbers in the table are the best case, and the underlined numbers in the table are the second-best case.
We compare different variants of our proposed methods and baseline with different numbers of noisy/corrupted cameras (ranging from $0$ to $2$).  We use a single-robot \textbf{baseline} with the same encoder and decoder structure. The baseline is trained on the clean dataset where the number of noisy cameras is $0$, and we test the baseline with all datasets with different number of noisy cameras. Our methods are trained and tested on the same datasets with all different numbers of noisy cameras. We also use a multi-robot \textbf{baseline-mp} which takes all the images as inputs and produces the desired outputs of all inputs. 
We study three different variants of our method: \textbf{mp-pose} represents multi-robot perception with spatial encoding messages, \textbf{mp-att} represents multi-robot perception with cross attention encoding messages, and \textbf{mp} represents multi-robot perception using messages without encoding, which are the previous-level node features. We use \textbf{mp} for ablation study to show the spatial and cross attention encoding effectiveness.
\vspace{-5pt}
\subsection{Qualitative Result}
\begin{figure}[!h]
    \centering
    \begin{subfigure}{0.22\linewidth}
    \includegraphics[width=\linewidth]{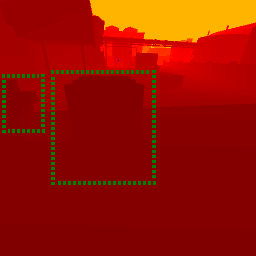}
    \end{subfigure}
    \begin{subfigure}{0.22\linewidth}
    \includegraphics[width=\linewidth]{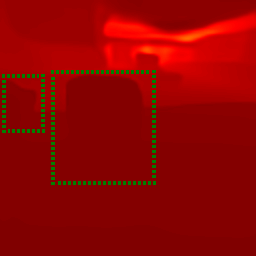}
    \end{subfigure}
    \begin{subfigure}{0.22\linewidth}
    \includegraphics[width=\linewidth]{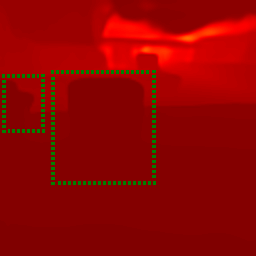}
    \end{subfigure}
    \begin{subfigure}{0.22\linewidth}
    \includegraphics[width=\linewidth]{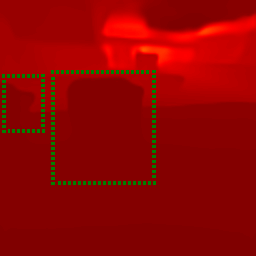}
    \end{subfigure}
    
    % Put a blank line here to divide into two rows
    \vspace{.6ex}
    \begin{subfigure}{0.22\linewidth}
    \includegraphics[width=\linewidth]{fig/fig1/10012_industrial_circle_baseline_0_depth_gt.png}
    \end{subfigure}
    \begin{subfigure}{0.22\linewidth}
    \includegraphics[width=\linewidth]{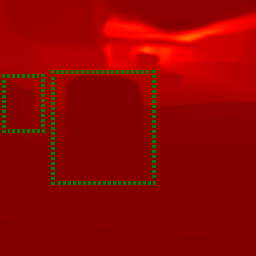}
    \end{subfigure}
    \begin{subfigure}{0.22\linewidth}
    \includegraphics[width=\linewidth]{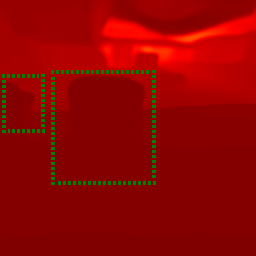}
    \end{subfigure}
    \begin{subfigure}{0.22\linewidth}
    \includegraphics[width=\linewidth]{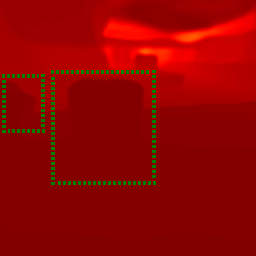}
    \end{subfigure}
    
    \vspace{.6ex}
    \begin{subfigure}{0.22\linewidth}
        \includegraphics[width=\linewidth]{fig/fig1/10012_industrial_circle_baseline_0_depth_gt.png}
    \end{subfigure}
    \begin{subfigure}{0.22\linewidth}
        \includegraphics[width=\linewidth]{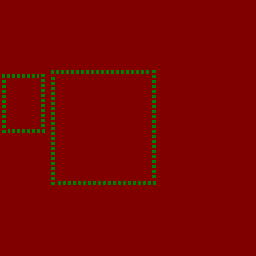}
    \end{subfigure}
    \begin{subfigure}{0.22\linewidth}
        \includegraphics[width=\linewidth]{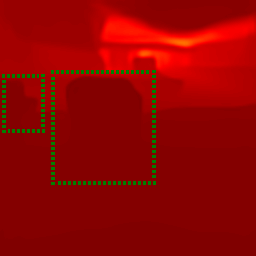}
    \end{subfigure}
    \begin{subfigure}{0.22\linewidth}
        \includegraphics[width=\linewidth]{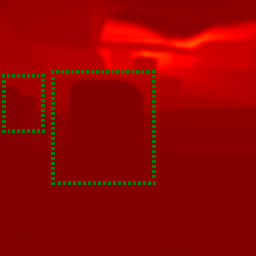}
    \end{subfigure}
    %\vspace{.6ex}
    \caption{Monocular depth estimation qualitative results varying the number of noisy cameras from a corrupted sensor. Methods from left to right: ground truth, baseline, mp, mp-pose. From top to bottom: 0, 1 and 2 noisy cameras.}\label{fig:5}
    \vspace{-15pt}
\end{figure}

\begin{figure}[!h]
    \centering
    \begin{subfigure}{0.22\linewidth}
        \includegraphics[width=\linewidth]{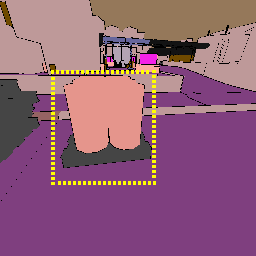}
    \end{subfigure}
    \begin{subfigure}{0.22\linewidth}
        \includegraphics[width=\linewidth]{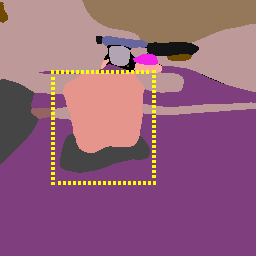}
    \end{subfigure}
    \begin{subfigure}{0.22\linewidth}
        \includegraphics[width=\linewidth]{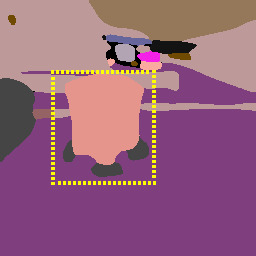}
    \end{subfigure}
    \begin{subfigure}{0.22\linewidth}
        \includegraphics[width=\linewidth]{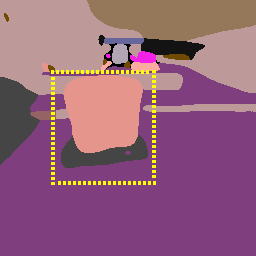}
    \end{subfigure}
    
    \vspace{.6ex}
    \begin{subfigure}{0.22\linewidth}
    \includegraphics[width=\linewidth]{fig/fig1/10012_industrial_circle_baseline_0_seg_seg_gt.png}
    \end{subfigure}
    \begin{subfigure}{0.22\linewidth}
    \includegraphics[width=\linewidth]{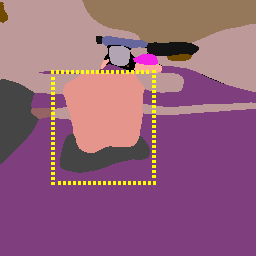}
    \end{subfigure}
    \begin{subfigure}{0.22\linewidth}
    \includegraphics[width=\linewidth]{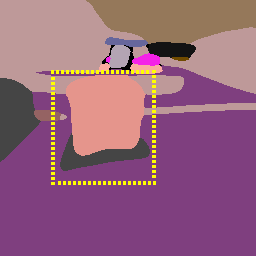}
    \end{subfigure}
    \begin{subfigure}{0.22\linewidth}
    \includegraphics[width=\linewidth]{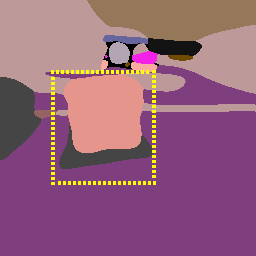}
    \end{subfigure}
    
    \vspace{.6ex}
    \begin{subfigure}{0.22\linewidth}
    \includegraphics[width=\linewidth]{fig/fig1/10012_industrial_circle_baseline_0_seg_seg_gt.png}
    \end{subfigure}
    \begin{subfigure}{0.22\linewidth}
    \includegraphics[width=\linewidth]{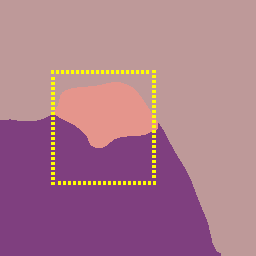}
    \end{subfigure}
    \begin{subfigure}{0.22\linewidth}
    \includegraphics[width=\linewidth]{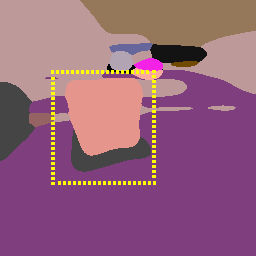}
    \end{subfigure}
    \begin{subfigure}{0.22\linewidth}
    \includegraphics[width=\linewidth]{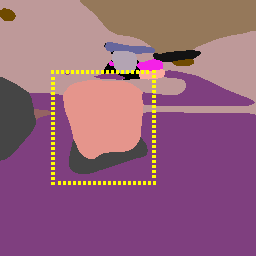}
    \end{subfigure}
    %\vspace{.6ex}
    \caption{Semantic segmentation qualitative results varying the number of noisy cameras. Methods from left to right: ground truth, baseline, mp, mp-pose. From top to bottom: 0, 1 and 2 noisy cameras.}\label{fig:6}
     \vspace{-10pt}
\end{figure}

\begin{figure}[!h]
    \centering
    \begin{subfigure}{0.15\linewidth}
    \includegraphics[width=\linewidth]{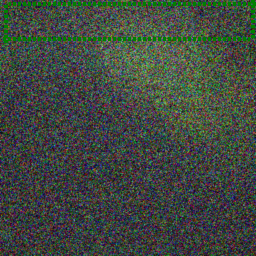}
    \end{subfigure}
    \begin{subfigure}{0.15\linewidth}
    \includegraphics[width=\linewidth]{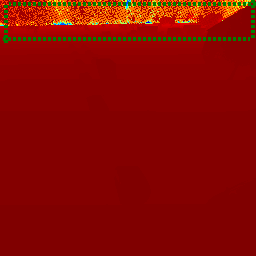}
    \end{subfigure}
    \begin{subfigure}{0.15\linewidth}
    \includegraphics[width=\linewidth]{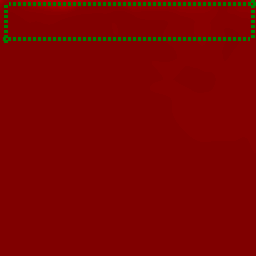}
    \end{subfigure}
    \begin{subfigure}{0.15\linewidth}
    \includegraphics[width=\linewidth]{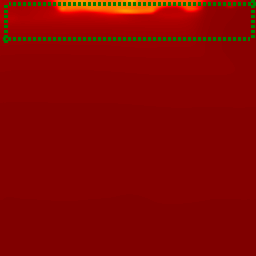}
    \end{subfigure}
    \begin{subfigure}{0.15\linewidth}
    \includegraphics[width=\linewidth]{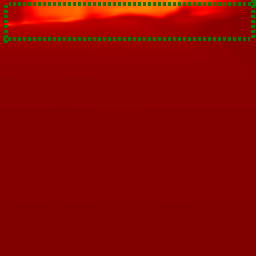}
    \end{subfigure}
    \begin{subfigure}{0.15\linewidth}
    \includegraphics[width=\linewidth]{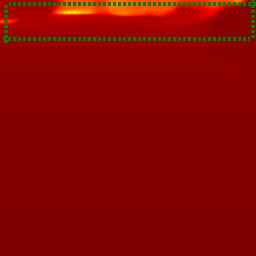}
    \end{subfigure}
    % Put a blank line here to divide into two rows
    
    \vspace{.6ex}
    \begin{subfigure}{0.15\linewidth}
    \includegraphics[width=\linewidth]{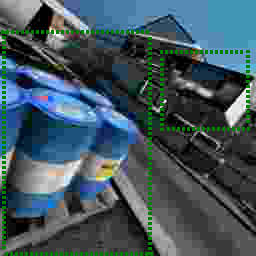}
    \end{subfigure}
    \begin{subfigure}{0.15\linewidth}
    \includegraphics[width=\linewidth]{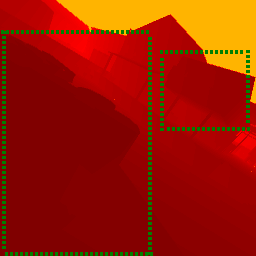}
    \end{subfigure}
    \begin{subfigure}{0.15\linewidth}
    \includegraphics[width=\linewidth]{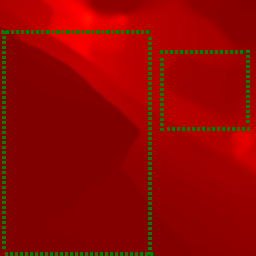}
    \end{subfigure}
    \begin{subfigure}{0.15\linewidth}
    \includegraphics[width=\linewidth]{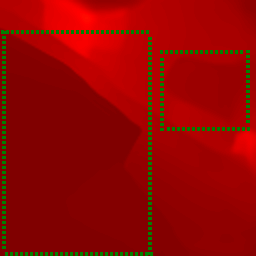}
    \end{subfigure}
    \begin{subfigure}{0.15\linewidth}
    \includegraphics[width=\linewidth]{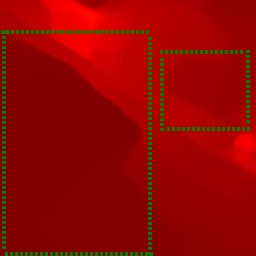}
    \end{subfigure}
    \begin{subfigure}{0.15\linewidth}
    \includegraphics[width=\linewidth]{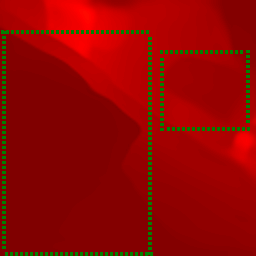}
    \end{subfigure}
    % Put a blank line here to divide into two rows
    
    \vspace{.6ex}
    \begin{subfigure}{0.15\linewidth}
    \includegraphics[width=\linewidth]{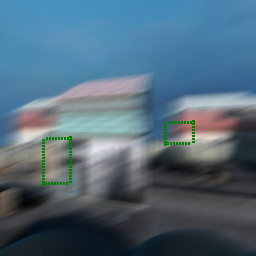}
    \end{subfigure}
    \begin{subfigure}{0.15\linewidth}
    \includegraphics[width=\linewidth]{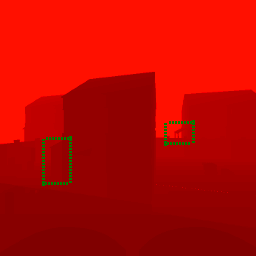}
    \end{subfigure}
    \begin{subfigure}{0.15\linewidth}
    \includegraphics[width=\linewidth]{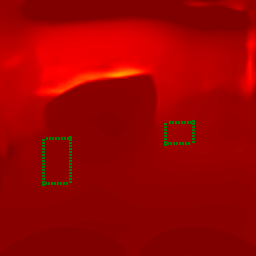}
    \end{subfigure}
    \begin{subfigure}{0.15\linewidth}
    \includegraphics[width=\linewidth]{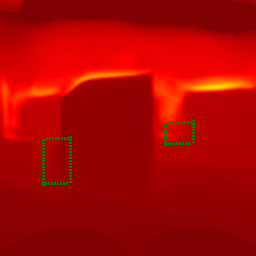}
    \end{subfigure}
    \begin{subfigure}{0.15\linewidth}
    \includegraphics[width=\linewidth]{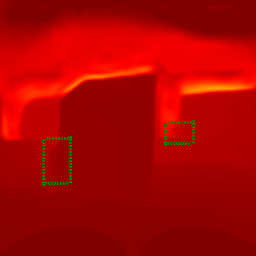}
    \end{subfigure}
    \begin{subfigure}{0.15\linewidth}
    \includegraphics[width=\linewidth]{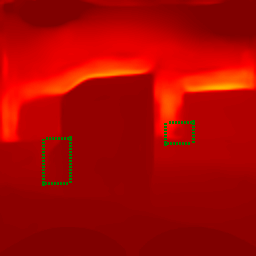}
    \end{subfigure}
    % Put a blank line here to divide into two rows
    %\vspace{.6ex}
    \caption{Monocular depth estimation qualitative results for different types of sensor noise. Methods from left to right: images, ground truth, baseline, mp, mp-pose, mp-att.}\label{fig:7}
    \vspace{-10pt}
\end{figure}

\begin{figure}[!h]
    \centering
    \begin{subfigure}{0.15\linewidth}
    \includegraphics[width=\linewidth]{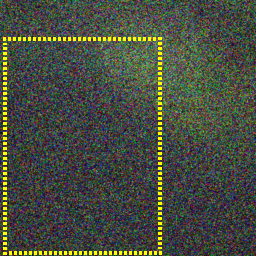}
    \end{subfigure}
    \begin{subfigure}{0.15\linewidth}
    \includegraphics[width=\linewidth]{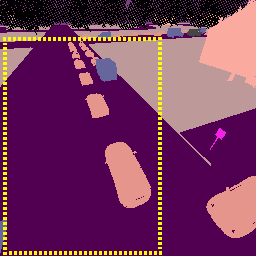}
    \end{subfigure}
    \begin{subfigure}{0.15\linewidth}
    \includegraphics[width=\linewidth]{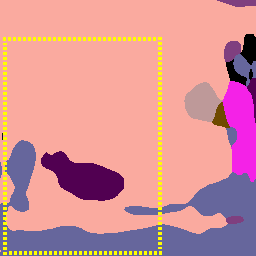}
    \end{subfigure}
    \begin{subfigure}{0.15\linewidth}
    \includegraphics[width=\linewidth]{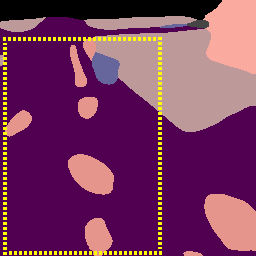}
    \end{subfigure}
    \begin{subfigure}{0.15\linewidth}
    \includegraphics[width=\linewidth]{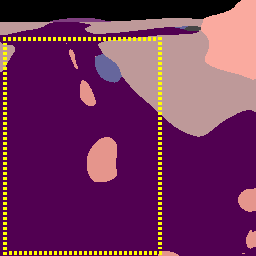}
    \end{subfigure}
    \begin{subfigure}{0.15\linewidth}
    \includegraphics[width=\linewidth]{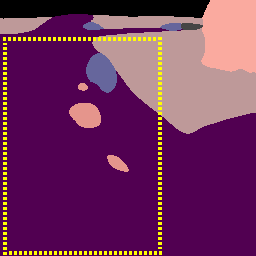}
    \end{subfigure}
    % Put a blank line here to divide into two rows
    
    \vspace{.6ex}
    \begin{subfigure}{0.15\linewidth}
    \includegraphics[width=\linewidth]{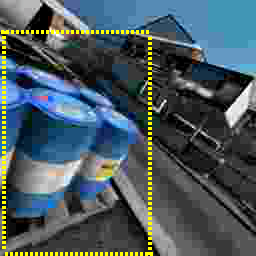}
    \end{subfigure}
    \begin{subfigure}{0.15\linewidth}
    \includegraphics[width=\linewidth]{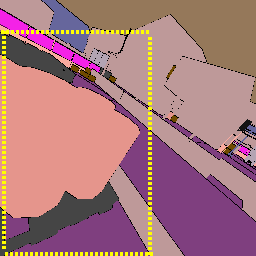}
    \end{subfigure}
    \begin{subfigure}{0.15\linewidth}
    \includegraphics[width=\linewidth]{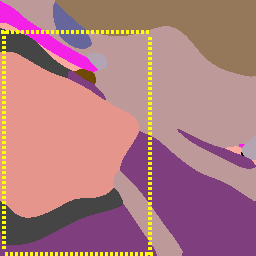}
    \end{subfigure}
    \begin{subfigure}{0.15\linewidth}
    \includegraphics[width=\linewidth]{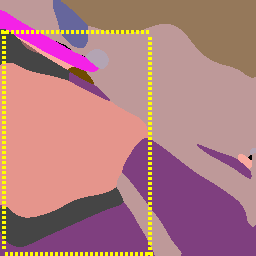}
    \end{subfigure}
    \begin{subfigure}{0.15\linewidth}
    \includegraphics[width=\linewidth]{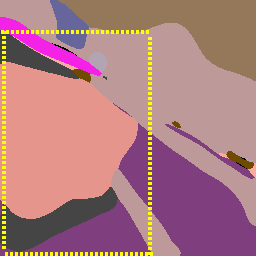}
    \end{subfigure}
    \begin{subfigure}{0.15\linewidth}
    \includegraphics[width=\linewidth]{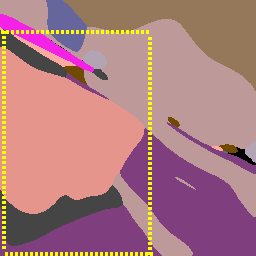}
    \end{subfigure}
    % Put a blank line here to divide into two rows
    
    \vspace{.6ex}
    \begin{subfigure}{0.15\linewidth}
    \includegraphics[width=\linewidth]{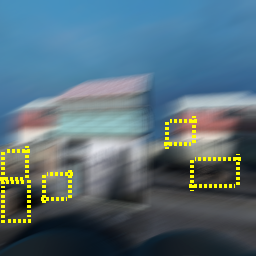}
    \end{subfigure}
    \begin{subfigure}{0.15\linewidth}
    \includegraphics[width=\linewidth]{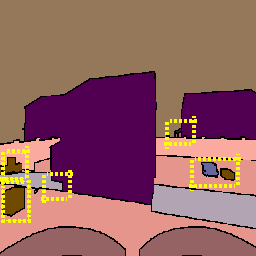}
    \end{subfigure}
    \begin{subfigure}{0.15\linewidth}
    \includegraphics[width=\linewidth]{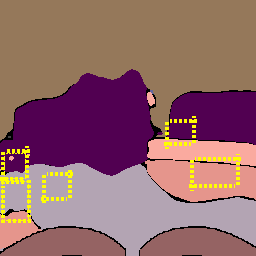}
    \end{subfigure}
    \begin{subfigure}{0.15\linewidth}
    \includegraphics[width=\linewidth]{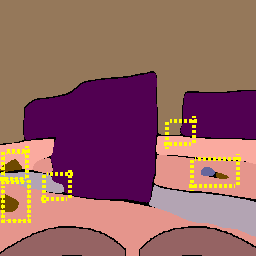}
    \end{subfigure}
    \begin{subfigure}{0.15\linewidth}
    \includegraphics[width=\linewidth]{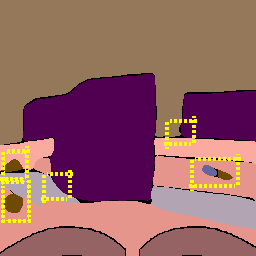}
    \end{subfigure}
    \begin{subfigure}{0.15\linewidth}
    \includegraphics[width=\linewidth]{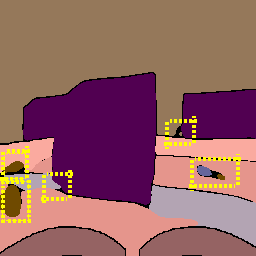}
    \end{subfigure}
    % Put a blank line here to divide into two rows
    %\vspace{.6ex}
    \caption{Qualitative results of semantic segmentation for different types of sensor noises. Methods from left to right:  images, ground truth, baseline, mp, mp-pose, mp-att.}\label{fig:8}
    \vspace{-20pt}
\end{figure}

In Fig.~\ref{fig:5} and Fig.~\ref{fig:6}, we show the result of \textbf{baseline}, \textbf{mp}, \textbf{mp-pose} with different number of noisy cameras. By increasing the number of noisy cameras, the performance of \textbf{baseline} decreases, and our methods: \textbf{mp-pose}, \textbf{mp-att} still preserve high-quality results. We can also see the effects of message encoding by comparing the details of the object boundaries and the perception accuracy of small and thin objects. In Fig.~\ref{fig:7} and Fig.~\ref{fig:8}, we use images from different datasets with different types of sensor corruption and noise. We highlight the area of object boundaries and small objects in the figures. Single-robot baseline creates wrong object boundaries and misses small and thin objects while our methods recover the information from corrupted images and preserve the perception details on both tasks.
These results show that the proposed method outperforms single-robot baseline and is robust to different types of image corruptions and noises.  This qualitative result shows that our method is robust to different numbers of noisy cameras in the robot network.

%%%%%%%%%%%%% 4*3

%%%%%%%%%%%%%%%%%%3*5 depth

%3*5 segmentation

\begin{table*}[ht]
    \centering
    \caption{Effects of spatial encoding and cross attention. Our method is robust to sensor noise.}
    \resizebox{\textwidth}{!}{%
    \begin{tabular}{cccccccccccccc}
    \toprule
         \multicolumn{2}{c}{\# noisy cameras} & \multicolumn{4}{c}{0} & \multicolumn{4}{c}{1} & \multicolumn{4}{c}{2} \\ \midrule
         dataset &	method &	Abs Rel$\downarrow$ &	Sq Rel$\downarrow$ &	RMSE$\downarrow$ &	mIoU$\uparrow$ & 	Abs Rel$\downarrow$ &	Sq Rel$\downarrow$ &	RMSE$\downarrow$ &	mIoU$\uparrow$ &	Abs Rel$\downarrow$&	Sq Rel$\downarrow$ &	RMSE$\downarrow$ &	mIoU$\uparrow$  \\ \midrule
         \multirow{5}{3em}{Airsim-MAP}& baseline & 0.0555 &	2.6438 &	18.6380 &	0.6170 &	0.3359 &	8.0912 &	37.1608 &	0.4154 & - & - & - & - \\
         & baseline-mp & 0.1556  & 6.7748	 & 36.2965	 & 0.3514 & 0.2489	& 18.1093 & 44.1963	& 0.2244	& - & - & - & - \\
         & mp-pose & \textbf{0.0521} &	\textbf{2.4841} &	\textbf{18.3737} &	\underline{0.6186} &	\textbf{0.0799} &	\textbf{4.3344} &	\textbf{25.4724} &	\underline{0.5055}  & - & - & - & -\\
         & mp-att & \underline{0.0542} &	\underline{2.5365} &	\underline{18.4385} &	\textbf{0.6251} &	\underline{0.0900} &	4.6000 &	\underline{25.9189} &	\textbf{0.5081}& - & - & - & -\\
         & mp & 0.0565 &	2.7646 &	18.6658 &	0.6151 &	0.0912 &	\underline{4.4830} &	26.7753 &	0.5033 & - & - & - & -\\ \midrule
         \multirow{5}{3em}{Industrial-pose}& baseline & 0.1290 &	{0.5918} &	3.8945 &	0.7718 &	0.2184 &	3.9593 &	13.6328	 & 0.6853 &	0.3085 &	7.5872 &	19.8783 &	0.6000 \\
         & baseline-mp & 0.3515   & 3.9684	  &	 10.5106  & 0.5096	  & 0.3793	 & 5.0247	 & 12.7207	 & 0.4903	 & 0.4147 & 6.5149 & 15.8023 & 0.4654 \\ 
         & mp-pose & \textbf{0.1174} &	\textbf{0.5608} &	\textbf{3.7336} &	\textbf{0.7808} &	\textbf{0.1251} &	{0.6564} & 	\underline{4.1251} &	{0.7446} &	\textbf{0.1277} &	\textbf{0.6276} &	\textbf{4.1556} &	\textbf{0.7436} \\
         & mp-att & \underline{0.1214}	& \underline{0.5817} &	\underline{3.7995} &	\underline{0.7800} &	\underline{0.1287} &	\textbf{0.5881} &	\textbf{4.1064} &	\textbf{0.7681}& 0.1385 &	0.6551&	4.2095&	\underline{0.7403}\\
         & mp & 0.1262 &	0.6032 &	{3.8019} &	{0.7791} &	{0.1294} &	\underline{0.6322} &	{4.1794} &	\underline{0.7469} &	\underline{0.1317} &	\underline{0.6360} &	\underline{4.2854} &	{0.7380}\\\bottomrule
    \end{tabular}
    }
    \label{tab:industrial-pose}
    \vspace{-5pt}
\end{table*}

\begin{table*}[t]
    \centering
    \caption{Study of the effect of
    sizes of overlapping areas across robots' FoV
    between robots. Larger overlap areas between robots FoV naturally provide more benefits from message passing with spatial encoding.}
    \resizebox{\textwidth}{!}{%
    \begin{tabular}{cccccccccccccc}
    \toprule
         \multicolumn{2}{c}{\# noisy cameras}& \multicolumn{4}{c}{0} & \multicolumn{4}{c}{1} & \multicolumn{4}{c}{2} \\ \midrule
         dataset &	method &	Abs Rel$\downarrow$ &	Sq Rel$\downarrow$ &	RMSE$\downarrow$ &	mIoU$\uparrow$ & 	Abs Rel$\downarrow$ &	Sq Rel$\downarrow$ &	RMSE$\downarrow$ &	mIoU$\uparrow$ &	Abs Rel$\downarrow$&	Sq Rel$\downarrow$ &	RMSE$\downarrow$ &	mIoU$\uparrow$  \\ \midrule
         \multirow{4}{3em}{Industrial-circle}& baseline & 0.0596 &	0.0995 &	\underline{0.7935}	& \underline{0.7623} &	0.1741 &	0.4763 &	1.9639 &	0.6895 &	0.2941 &	0.8912 &	2.7556 &	0.6144 \\
         & baseline-mp & 0.4376 & 1.4737	& 2.4441	&  0.5397 &  0.5530 & 2.4935	& 2.7593	& 0.5132	& 0.7293	& 4.1689	& 3.3767	 & 0.4886 \\
         & mp-pose & \textbf{0.0576} &	\underline{0.0930} &	\textbf{0.7880} &	\textbf{0.7662} &	\textbf{0.0827} &	\textbf{0.1359} &	\textbf{0.8923} &	\textbf{0.7362} & 	\textbf{0.1269} &	\underline{0.2934} &	\textbf{0.9288} &	\textbf{0.7340} \\
         & mp & \underline{0.0582} &	\textbf{0.0914} &	0.7939 &	0.7597 &	\underline{0.0851} &	\underline{0.1564} &	\underline{0.9111} &	\underline{0.7355} &	\underline{0.1314} &	\textbf{0.2714} &	\underline{0.9415} &	\underline{0.7220}\\\midrule
         \multirow{4}{3em}{Industrial-rotation}& baseline & \textbf{0.1107} &	\underline{0.1340} &	\underline{0.7338} &	0.7631 &	0.3182 &	0.5354 &	1.5238 &	0.7107 &	0.4531 &	0.8693 &	2.0525 &	0.6618 \\
         & baseline-mp &  0.4761 & 0.8329	& 1.8128	& 0.6524 & 0.7458	 & 1.3347 & 2.0131 & 0.6259		& 1.1000 &	2.0564 & 2.3251 & 0.5985	\\
         & mp-pose  & 0.1173 &	\textbf{0.1259}&	\textbf{0.7295} &	\textbf{0.7657} & 	\underline{0.2228} &	\underline{0.4834} &	\textbf{0.9540} &	\textbf{0.7387} & 	\textbf{0.2277} &	\underline{0.2918} &	\textbf{1.0274} &	\textbf{0.7276} \\
         & mp & \underline{0.1157} &	0.1349 &	0.7432 &	\underline{0.7639} &	\textbf{0.2115} &	\textbf{0.2893} &	\underline{0.9812} &	\underline{0.7368} &	\underline{0.2307} &	\textbf{0.2706} &	\underline{1.0404} &	\underline{0.7212}\\\bottomrule
    \end{tabular}
    }
    \label{tab:formation}
    \vspace{-5pt}
\end{table*}
\begin{table*}[!h]
    \centering
    \caption{Real-world monocular depth estimation experiments. Our methods outperform the baseline on both tasks.}
    \begin{tabular}{ccccccccccc}
    \toprule 
         \multicolumn{2}{c}{\# noisy cameras}& \multicolumn{3}{c}{0} & \multicolumn{3}{c}{1} & \multicolumn{3}{c}{2} \\ \midrule
         dataset &	method &	Abs Rel$\downarrow$ &	Sq Rel$\downarrow$ &	RMSE$\downarrow$ &		Abs Rel$\downarrow$ &	Sq Rel$\downarrow$ &	RMSE$\downarrow$ &	Abs Rel$\downarrow$&	Sq Rel$\downarrow$ &	RMSE$\downarrow$ \\ \midrule
         \multirow{3}{3em}{ARPL-r4}& baseline & \underline{0.1604} &	\underline{0.3416} &	\underline{0.4221}  &	0.1886 &	0.3812 &	0.5480  &	0.2194 &	0.4206 &	0.6591  \\
         & mp-pose & \textbf{0.1566} &	\textbf{0.3370} &	\textbf{0.4079}  &	\underline{0.1822} &	\underline{0.3540} &	\textbf{0.4479}  &	\textbf{0.1700} &	\textbf{0.3454} &	\textbf{0.4351}  \\
         & mp & 0.1844 &	0.3562 &	0.4614  &	\textbf{0.1797} &	\textbf{0.3516} &	\underline{0.4668}  &	\underline{0.1999} &	\underline{0.3718} &	\underline{0.4955} \\    \bottomrule
    \end{tabular}
    \label{tab:arpl}
    \vspace{-10pt}
\end{table*}
\vspace{-5pt}
\subsection{Quantitative Result}

\subsubsection{Effects of message encoding}\label{sec:effect-message-encoding}
To show the the effects of the message passing with spatial encoding, we use \textbf{Airsim-MAP} dataset with sensor corruption and \textbf{Industrial-pose} dataset with sensor noises.  
In Table~\ref{tab:industrial-pose}, we observe that \textbf{mp-pose} outperforms \textbf{baseline} on both tasks among all datasets with different number of noisy cameras. Comparing \textbf{mp-pose} and \textbf{mp}, we find that the spatial encoding improves the performance of both tasks in most of the experiments.
We also demonstrate the effects of the cross attention message encoding on \textbf{Airsim-MAP} dataset and \textbf{Industrial-pose} dataset. In Table~\ref{tab:industrial-pose}, we show that \textbf{mp-att} outperforms \textbf{baseline} on all cases in both tasks, and \textbf{mp-att} outperforms \textbf{mp} on most cases in both tasks. 
We also demonstrates the necessity of GNN in the multi-robot perception task by comparing \textbf{baseline-mp} with other methods, which shows that \textbf{baseline-mp} are outperformed by all other methods, since it cannot learn the correspondence of features by message passing. 
We can conclude that both message encoding including the spatial encoding and the cross attention encoding improves the accuracy of both tasks, and its robustness with respect to different numbers of corrupted and noisy image cameras.

\subsubsection{Effects of overlap between robots}
In Table~\ref{tab:formation}, we study how the overlapping areas between robots' FoV affect the proposed message passing with spatial encoding. We use \textbf{Industrial-circle} dataset and \textbf{Industrial-rotation} for this study. We find that our method \textbf{mp-pose} outperforms \textbf{baseline} for both datasets, which indicates that our method is robust to different sizes of overlapping areas across robots' FoV in the robot network. Comparing the performance gap between \textbf{mp-pose} and \textbf{mp} on both datasets, we can find that this is smaller for \textbf{Industrial-circle} dataset rather than \textbf{Industrial-rotation}, especially for the mIoU metric in the semantic segmentation case. This result shows that the message passing with spatial encoding can be affected by the overlapping pattern. Therefore, for  limited overlapping areas, we correctly obtain reduced benefits from the message passing mechanism. 

\subsubsection{Real-world experiments}
We demonstrate our approach on a real-world dataset \textbf{ARPL-r4}. We use onboard visual-inertial odometry to calculate the relative pose between robots. Since we have access to the relative spatial relationship between robots, we use messages with spatial encoding. In Table~\ref{tab:arpl}, we show that our method \textbf{mp-pose} with spatial encoded message outperforms the baseline as well as the one without spatial encoding in most cases. 

\vspace{-5pt}
\subsection{Discussion}
In the following, we discuss the bandwidth requirements for the proposed and the corresponding benefits compared to a direct exchange of sensor data among the robots as well as the advantages and disadvantages for the two encoding mechanisms. 
We employ MBytes per frame (MBpf) as the metric to measure the bandwidth requirements of our approach. By learning and communicating encoded messages, the communication bandwidth for one message between two robots for one level is $2.5$ Mbpf, while the communication bandwidth of directly sharing raw sensor measurements is $6$ Mbpf. The levels of the GNN are usually limited to prevent over-smoothing of the GNN. In the demonstrated case, we do not have more than $2$ levels which makes the approach communication-efficient than sharing the original information.
Also, the message passing can reduce the computation requirements by spreading the computations to each agent, which makes the computation requirements feasible for the robot team.
This further shows the ability to scale the proposed approach to larger teams. It is possible to further reduce communication bandwidth by adopting smaller network encoder or learning communication groups as shown in~\cite{liu2020When2com}.
For other sensing modalities, similar bandwidth trade-offs are expected.

The bandwidth requirements is equivalent for the two proposed encoding mechanisms, but these have different properties. The results in Section~\ref{sec:exepriements} confirm that the spatial encoding can utilize relative poses when this information is accessible, but it cannot dynamically capture the correlation between the sensor measurements. Conversely, the dynamic cross attention encoding can utilize the correlation relationship among features when the relative pose is not accessible. Geometric-aware tasks like monocular depth estimation benefits more by spatial encoding since the relative poses contain important geometric cues,  while semantic-aware tasks like semantic segmentation do not favor either one of the mechanism.
In the future, it would be interesting to investigate how to combine the advantages of both mechanisms.

%% file: chap/5_conclusion.tex
In this work, we proposed a multi-robot collaborative perception framework with graph neural networks. We utilize message passing mechanism with spatial encoding and cross attention encoding to enable information sharing and fusion among the robot team. The proposed method is a general framework which can be applied to different sensor modalities and tasks. We validate the proposed method on semantic segmentation and monocular depth estimation tasks on simulated  and real-world datasets collected from multiple aerial robots' viewpoints including different types of severe sensor corruptions and noises. 
The experiment shows that the proposed method increases the perception accuracy and robustness to sensor corruption and noises. It can be an effective solution for different multi-view tasks with swarms of aerial or ground robots.

In the future, we will extend this framework to other sensor modalities and tasks. We will also explore how to use the graph neural network to integrate perception, control and planning framework for aerial swarm systems.